\documentclass[preprint,12pt,authoryear]{elsarticle}

\usepackage{amssymb}
\usepackage[]{geometry}
\usepackage{changepage}
\usepackage{tabularray}
\usepackage{booktabs}
\usepackage{caption} 
\usepackage{amsmath}
\usepackage{lscape}
\usepackage{multirow}
 \usepackage{graphicx}
\journal{preprint}

\begin{document}

\begin{frontmatter}



\title{RADARSAT Constellation Mission Compact Polarisation SAR Data for Burned Area Mapping with Deep Learning} 

\cortext[cor1]{Corresponding author}
\author{Yu Zhao, Yifang Ban\corref{cor1}} 

\affiliation{organization={KTH Royal Institute of Technology},
            addressline={Tekniskringen 10A}, 
            postcode={10031}, 
            state={Stockholm},
            country={Sweden}}

\begin{abstract}
Monitoring wildfires has become increasingly critical due to the sharp rise in wildfire incidents in recent years. Optical satellites like Sentinel-2 and Landsat are extensively utilized for mapping burned areas. However, the effectiveness of optical sensors is compromised by clouds and smoke, which obstruct the detection of burned areas. Thus, satellites equipped with Synthetic Aperture Radar (SAR), such as dual-polarization Sentinel-1 and quad-polarization RADARSAT-1/-2 C-band SAR, which can penetrate clouds and smoke, are investigated for mapping burned areas. However, there is limited research on using compact polarisation (compact-pol) C-band RADARSAT Constellation Mission (RCM) SAR data for this purpose. This study aims to investigate the capacity of compact polarisation RCM data for burned area mapping through deep learning. Compact-pol m-$\chi$ decomposition and Compact-pol Radar Vegetation Index (CpRVI) are derived from the RCM Multi-look Complex product. A deep-learning-based processing pipeline incorporating ConvNet-based and Transformer-based models is applied for burned area mapping, with three different input settings: using only log-ratio dual-polarization intensity images images, using only compact-pol decomposition plus CpRVI, and using all three data sources. The training dataset comprises 46,295 patches, totalling 90 GB, generated from 12 major wildfire events in Canada. The test dataset includes seven wildfire events from the 2023 and 2024 Canadian wildfire seasons in Alberta, British Columbia, Quebec and the Northwest Territories. The results demonstrate that compact-pol m-$\chi$ decomposition and CpRVI images significantly complement log-ratio images for burned area mapping. The best-performing Transformer-based model, UNETR, trained with log-ratio, m-$\chi$ decomposition, and CpRVI data, achieved an F1 Score of 0.718 and an IoU Score of 0.565, showing a notable improvement compared to the same model trained using only log-ratio images (F1 Score: 0.684, IoU Score: 0.557).
\end{abstract}


\begin{highlights}

\item 1) Demonstrating the compact-pol m-$\chi$ decomposition and compact-pol vegetation index images, generated from RCM C-band Multilook Complex product, complement the log-ratio intensity images effectively for burned area mapping.

\item 2) Combining compact-pol RCM data with ConvNet-based and Transformer-based segmentation models, the proposed workflow can effectively detect wildfires in 7 study regions at 9 timestamps in the test dataset. The Transformer-based model UNETR provides the best F1 Score of 0.718 and IoU Score of 0.565 among other baseline models.
\end{highlights}


\end{frontmatter}



\section{Introduction}
\label{intro}
Wildfires have become a major natural hazard, with sharp increases in both intensity and frequency \cite{Cunningham2024IncreasingFA}. During the 2023 wildfire season, Canada endured 6,694 wildfires, resulting in 18.5 million hectares burned—the largest area burned in Canada’s recorded history \cite{fire}. Similarly, in 2023, Greece recorded a total of 80 wildfires including one of the largest wildfires in European Union history, with over 73,000 hectares burned \cite{a2023_firefighters}. Burned area mapping is a crucial for wildfire management, providing key insights into the progression of ongoing wildfires and their impact on the landscape and ecosystems. \par  
Earth observation data are widely used in burned area mapping for cost-effectiveness on a large scale. Optical sensors like Sentinel-2 MultiSpectral Instrument (MSI) or Landsat Operational Land Imager (OLI) are commonly used in mapping burned areas \cite{Seydi2022BurntNetWB, Roy2019Landsat8AS}. Both Sentinel-2 MSI and Landsat OLI provide images at 10-30 meters of medium spatial resolution, making them suitable for burned area mapping. However, for optical sensors, the most challenging issue is cloud interference. Capable of penetrating cloud cover, satellites equipped with SAR sensor is an alternative approach to map burned areas and monitor wildfire progression \cite{ban2020near, Zhang2023TotalvariationRU}. The backscatter in SAR images changes between pre-fire and post-fire conditions due to alterations in the ground biomass structure caused by wildfires. SAR's effectiveness in detecting burned areas varies with different wavelengths \cite{ban2020near, Zhao2022GlobalSB}. C-band SAR is proficient in detecting changes in volume scattering from tree canopies and branches. \cite{Luft2022DeepLB} uses U-Net and Sentinel-1 C-band data to detect burned areas in California. \cite{BelenguerPlomer2021CNNbasedBA} proposes to use the fusion of optical and radar data to improve detection accuracy. A study by \cite{ban2020near} proposed a near real-time wildfire progression mapping method using deep learning and C-band Sentinel-1 Data, showcasing the potential of C-band SAR for mapping burned areas. Furthermore, \cite{Zhang2021LearningUW} demonstrated that continuously fine-tuning a model can enhance its performance in identifying new wildfire sites. This body of research underscores the diverse capabilities of SAR at different wavelengths for effective wildfire monitoring. Launched in 2019, the RADARSAT Constellation Mission (RCM) operates at the same wavelength as Sentinel-1 and comprises three identical satellites, enabling more frequent revisits. Different from the dual polarisation Sentinel-1 provides, RCM provides compact polarisation that enables circularly transmitting signals and receiving Horizontally and Vertically. Compact polarisation can collect more scatter information of the target object while remaining the simple design of dual polarisation sensors \cite{White2017MovingTT}. Since the circular signal combines horizontal and vertical transmitting signals, compact polarisation can obtain comparable scattering information to quad polarisation \cite{Chen2020ANS}. While research specifically focusing on using RCM for burned area mapping is limited, there is a substantial body of work investigating the use of previous RADARSAT satellites for this purpose. The study by \cite{goodenough2011mapping} demonstrates a method that utilizes RADARSAT-2 data in conjunction with the k-means algorithm to map burned areas. \cite{gimeno2004evaluation} presents an analysis of RADARSAT-1 data in detecting burned scars under different incidence angles and shows a method based on a neural network for burned area mapping. While these previous studies laid a solid foundation for SAR-based burned area mapping, further research is needed to investigate the RCM compact polarization SAR for burned area mapping. For burned area mapping using RCM compact-pol data, a recent study \cite{Chen2023DerivationAA} indicates the Radar Forest Degradation Index is effective in burned area mapping. The objective of this work is to investigate the capacity of compact polarisation RCM data for burned area mapping when utilizing deep-learning-based methods. The major contributions can be summarized in two folds:
\begin{itemize}
    \item Demonstrating the compact-pol m-$\chi$ decomposition and compact-pol vegetation index images, generated from RCM C-band Multilook Complex product, complement the log-ratio intensity images effectively for burned area mapping.
    \item Combining compact-pol RCM data with ConvNet-based and Transformer-based segmentation models, the proposed workflow can effectively detect wildfires in 7 study regions at 9 timestamps in the test dataset. The Transformer-based model UNETR provides the best F1 Score of 0.718 and IoU Score of 0.565 among other baseline models.
\end{itemize}
\section{Related Studies}
\subsection{Applications of RCM compact-pol data}
 Related studies on other applications utilizing RCM compact-pol data are covered in this section to provide insights into the usage of compact-pol data. \cite{DingleRobertson2022MonitoringCU} uses compact-pol data and the m-$\chi$ decomposition data for operational crop classification in Canada. The promising results generated by a random forest classifier show that m-$\chi$ data can achieve equivalent results compared to optical data. \cite{Mahdianpari2019MidseasonCC} also utilizes random forest algorithm and RCM compact-pol data for crop classification. Although, it compares the performance of using compact-pol data with using full polarisation and dual polarisation. The results indicate that compact-pol can achieve significantly better classification accuracy compared to dual polarization. However, compared with full polarisation data, there is still a performance gap with compact-pol data. The application of crop classification highlights the possibility of compact-pol decomposition data being used for vegetation classification. RCM data can also be used for soil moisture retrieval\cite{Dabboor2024TheRC}. The soil moisture retrieval algorithm proposed by \cite{Dabboor2024TheRC} achieves a high correlation with R-square equals 0.75 compared to in-situ measurement. Similarly, RCM compact-pol is also used for peatland classification \cite{White2017MovingTT}. The result is also compared with full-polarisation Radarsat-2 data, which shows comparable accuracy. For forestry applications, \cite{Chen2020ANS} compared the radar vegetation index generated from compact-pol (RCM) and quad-pol (Radarsat-2) data. It shows that the radar vegetation index of compact-pol shows strong agreement with the index generated from qual-pol for different landcovers. This promising result encourages the usage of radar vegetation index from compact-pol data for forestry applications.\par

In summary, among all applications, the results derived from RCM compact-pol data demonstrate substantial improvements over dual-polarization data. When compared to full-polarization data, the performance of compact-pol data is comparable but still has a performance gap compared to full-polarization results. Additionally, features such as compact-pol decomposition and the radar vegetation index have proven their importance in various applications, motivating the utilization of these features in burned area mapping.
\section{Deep Learning Segmentation Models for Wildfire Detection}
\subsection{ConvNet-based Models}
Semantic segmentation is a computer vision task that seeks to classify every pixel in an image into specific categories. Deep learning models have become widely used for this task, with Convolutional Neural Networks (ConvNets) being the most popular choice. In recent years, ConvNet-based segmentation models have also gained traction in Earth observation applications, demonstrating their effectiveness in tasks like wildfire monitoring. The Fully Convolutional Network (FCN) was the first segmentation model to utilize only ConvNet, achieving pixel-wise predictions with a deep ConvNet encoder and a lightweight decoder. U-Net improved upon FCN by balancing the encoder and decoder architectures \cite{unet}. It also enabled more detailed segmentation by propagating high-level texture information from the encoder to the decoder. Attention U-Net further enhanced U-Net by introducing an attention mechanism in the encoder \cite{Oktay2018AttentionUL}, helping to focus the model on relevant parts of the image for more accurate segmentation.\par
As for wildfire detection applications, semantic segmentation models are commonly used for detecting the active fire and the burned area. In \cite{ban2020near} and \cite{Zhang2023TotalvariationRU}, U-Net is used to monitor the burned area using Sentinel-1 C-band SAR imagery. It also utilizes the pseudo masks generated from the image statistics to train the model, which shows promising results in mapping the burned area. \cite{Zhang2023TotalvariationRU} further improves the deep learning workflow to include total variation loss which helps the model to generate fine-grained burned area maps. For L-band SAR data, \cite{Zhao2022GlobalSB} tests six different ConvNet-based models for burned area mapping. The result indicates that U-Net is the most efficient in detecting burned areas using L-band SAR data compared to other baseline deep learning models. For optical data like Sentinel-2, U-Net is also proven to be effective in detecting burned areas \cite{knopp2020deep}.
\subsection{Transformer-based Models}
The Transformer model is proposed for Natural Language Processing. The self-attention and cross-attention module introduced shows significant improvement in generating latent representations of text\cite{vaswani2017attention}. By adopting the Transformer model for computer vision tasks, Vision Transformer \cite{kolesnikov2021image} is the first Transformer-based model used for semantic segmentation and image classification. To solve the scaling issue when the size of the images increases, Swin-Transformer \cite{Liu2021SwinTH} is introduced by improving the efficiency in the attention module by introducing the attention window mechanism. For Transformer-based models, the self-attention mechanism helps to generate strong representations of images\cite{Lin2017ASS}. For semantic segmentation tasks, Transformer-based segmentation models become increasingly popular after the success of the Vision Transformer. Recently, UNETR \cite{hatamizadeh2022unetr} and SwinUNETR/SwinUNETR-V2 \cite{Tang2021SelfSupervisedPO, He2023SwinUNETRV2SS} have emerged as strong semantic segmentation baselines, designed to handle three-dimensional images. UNETR and SwinUNETR utilize Vision Transformer and Swin-Transformer encoders to enhance segmentation performance.\par
For wildfire detection tasks using Transformer-based models, \cite{Han2024BurnedAA} applies a Transformer-based change detection model for burned area mapping and burn severity mapping. It indicates that Transformer-based models can achieve promising accuracy in both fire detection tasks. \cite{Gonalves2023TransformersFM} compared the performance of Transformer-based models and ConvNet-based models for burned area mapping in the Amazon forest with optical PlanetScope image, proving Transformer-based models show privilege in detecting burned areas. \cite{Gerard2023WildfireSpreadTSAD} also compared the performance of Transformer-based models with ConvNet-based models for wildfire progression prediction tasks with optical data, showing that Transformer-based models can improve the prediction accuracy from ConvNet-based models. \cite{Zhao2023TokenizedTI} shows that the Transformer-based model also performs better than ConvNet-based models in active fire detection.
\section{RCM Dataset Description}
\subsection{Study Areas}
\begin{figure*}[hbt!]
  \centering
  \includegraphics[width=\linewidth]{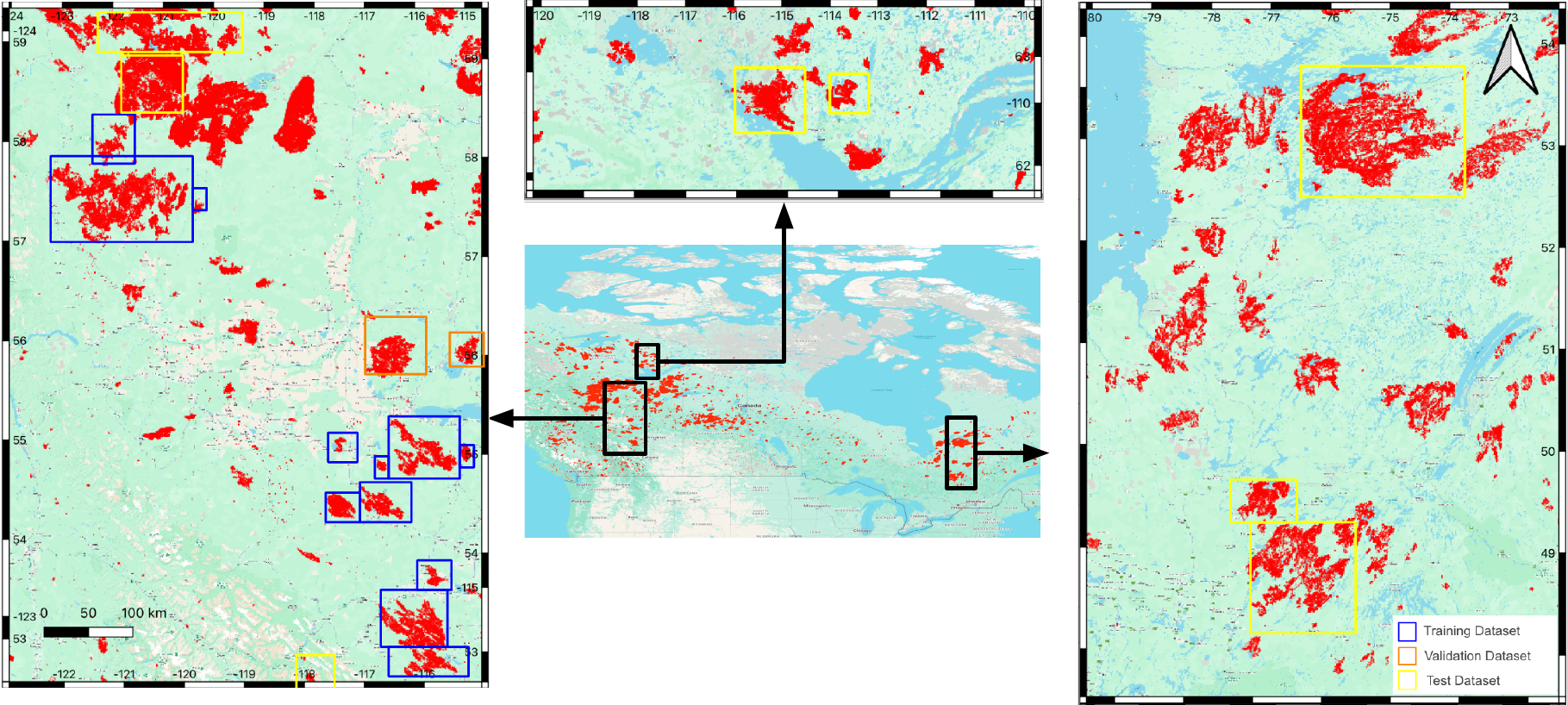}
\caption{Selected wildfire events used in training, validation and test dataset from British Columbia, Alberta and Quebec, Canada in 2023 and 2024.}
\label{fig:sa}
\end{figure*}

As shown in Figure \ref{fig:sa}, the dataset is collected based on wildfire events in Alberta, British Columbia, Quebec and the Northwest Territory, Canada in 2023 and 2024. The burned areas of the wildfire events are shown using the  Active Fire data (VNP14IMG) from NASA Visible Infrared Imaging Radiometer Suite (VIIRS) images.\par
In total 12 wildfire events are selected to generate a training dataset with 46295 image patches and two wildfire events are used as the validation dataset with 2976 image patches. Each image patch has a size of 256 by 256. The pre- and post-fire image pairs with the same incidence angles within the period from May 1st to October 31st, 2023 are collected as the training set and test set. Consequently, a training dataset consisting of pre- and post-fire image pairs with a size of 90GB is created. For the test set, seven Canadian wildfire events with a total of nine sets of pre and post-fire images with the same incidence angles are used, as described in detail in Section \ref{quantitative}. Two of these wildfire events, located in Quebec, are collected and evaluated together due to their close geolocation.\par 

\subsection{Dataset Description}
\begin{figure}[!hbt]
    \centering
    \includegraphics[width=.7\linewidth]{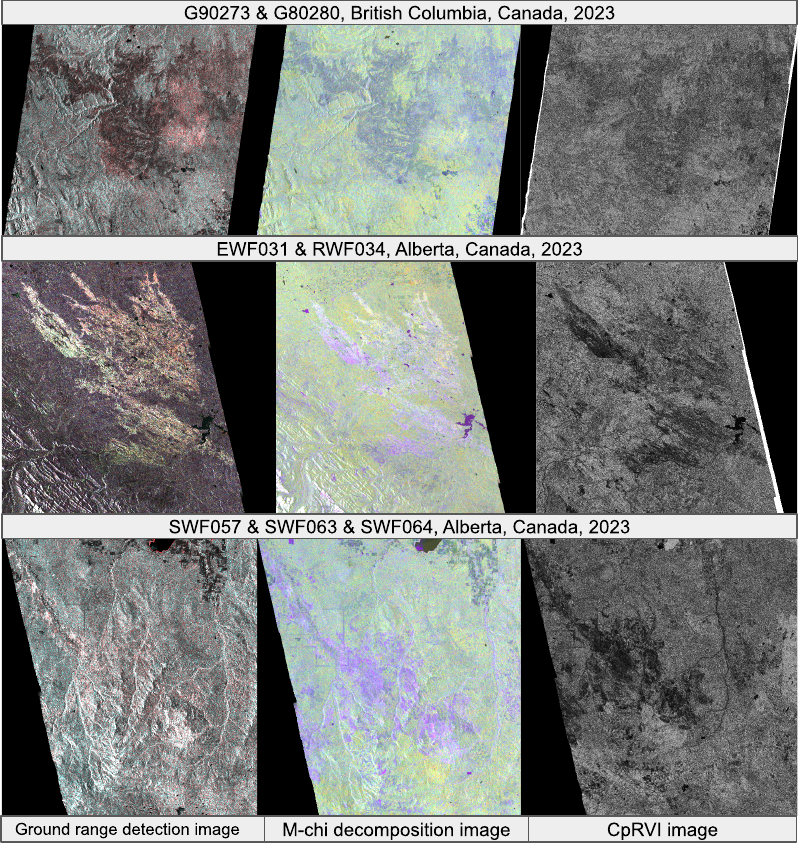}
    \caption{Visualization of GRD image, m-$\chi$ decomposition image and CpRVI image for three wildfire events of the training dataset.}
    \label{fig:inspect}
\end{figure}

RADARSAT Constellation Mission consists of three identical satellites with C-band SAR sensors at 5.405 GHz frequency. The Multi Look Complex (MLC) products captured in ScanSAR mode at a spatial resolution of 30 meters are used which provides the best spatial resolution. There are four different beam modes available with beam mnemonic (SC30MCP[A-D]) which represent different incidence angles as shown in Figure \ref{tab:beammode}. Both ascending and descending orbits are utilized in the MLC products. The compact-pol Multi-Look Complex products store a two by two matrix as shown in Equation \ref{c2matrix}. The diagonal elements are real number elements which are the squares of two compact polarisation channels. The off-diagonal elements are complex number elements which are the products of one channel and the conjugate of the other channel. 
\begin{equation}
\label{c2matrix}
    C2=\begin{bmatrix}
|CH|^2& CH\cdot CV^*\\
CV\cdot CH^* & |CV|^2
\end{bmatrix}
\end{equation}
As shown in the first column of Figure \ref{fig:inspect}, the ground range detection (GRD) images derived from the MLC products are presented. GRD images are derived from MLC products after terrain correction. They have two polarisation bands CH and CV, which represent the circular signals from the transmitter received by horizontal (CH) and vertical antennas (CV), and one intensity band. The detail of the processing pipeline is described in Section \ref{pre-proc}. The burned area within the GRD images shows a different pattern among the three examples in Figure \ref{fig:inspect}. As fire removes vegetation, leaving bare soil and ash, burned areas usually have lower radar backscatter compared to unburned vegetation, leading to a darker colour than the unburned areas. For the other two areas, the burned areas are brighter than the unburned areas due to the increase in backscatter, likely caused by precipitation.

\begin{table}[!hbt]
\centering
\caption{Incidence angles of four beam modes SC30MCP[A-D]}
\label{tab:beammode}
\scalebox{0.8}{
\begin{tblr}{
  cells = {c},
  hline{1-2,6} = {-}{},
}
{\textbf{Beam Mode }\\\textbf{Mnemonic}} & {\textbf{Min Incidence }\\\textbf{Angle}} & {\textbf{Max Incidence }\\\textbf{Angle}} \\
SC30MCPA                                 & 17.30                                     & 28.84                                     \\
SC30MCPB                                 & 26.09                                     & 36.30                                     \\
SC30MCPC                                 & 33.89                                     & 42.75                                     \\
SC30MCPD                                 & 40.67                                     & 48.30                                     
\end{tblr}}
\end{table}
\subsubsection{Compact polarisation Decomposition}
\begin{figure}[hbt!]
  \centering
  \includegraphics[width=\linewidth]{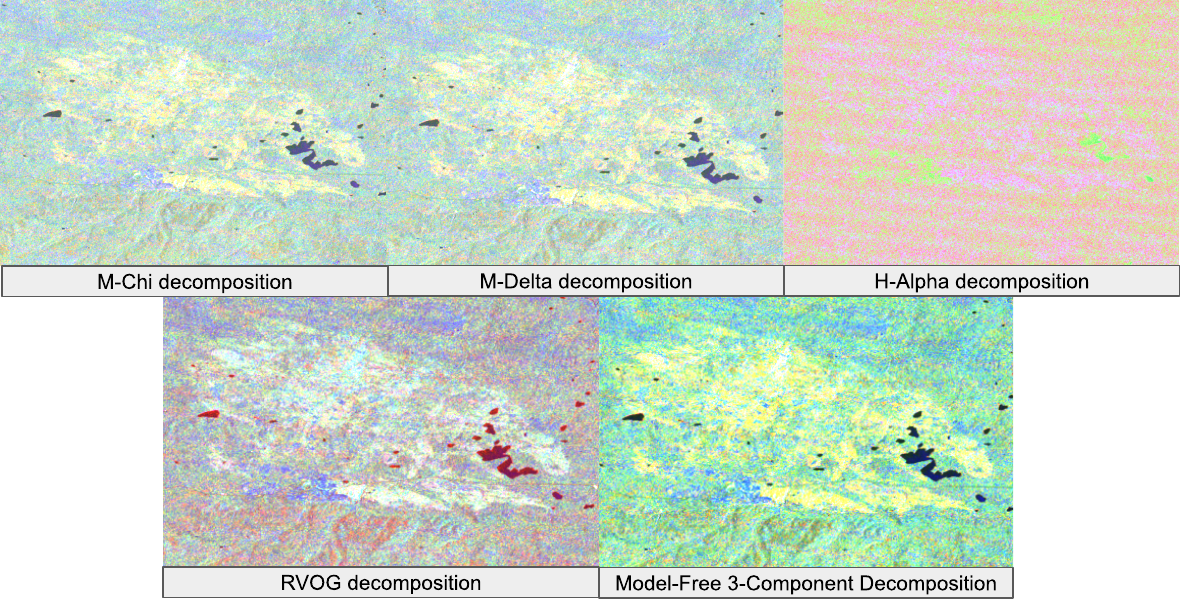}
\caption{Comparison between images generated from 5 compact polarization decomposition method}
\label{fig:decomp_comp}
\end{figure}

\begin{figure}[hbt!]
  \centering
  \fontsize{6}{6}
  \includegraphics[scale=0.55]{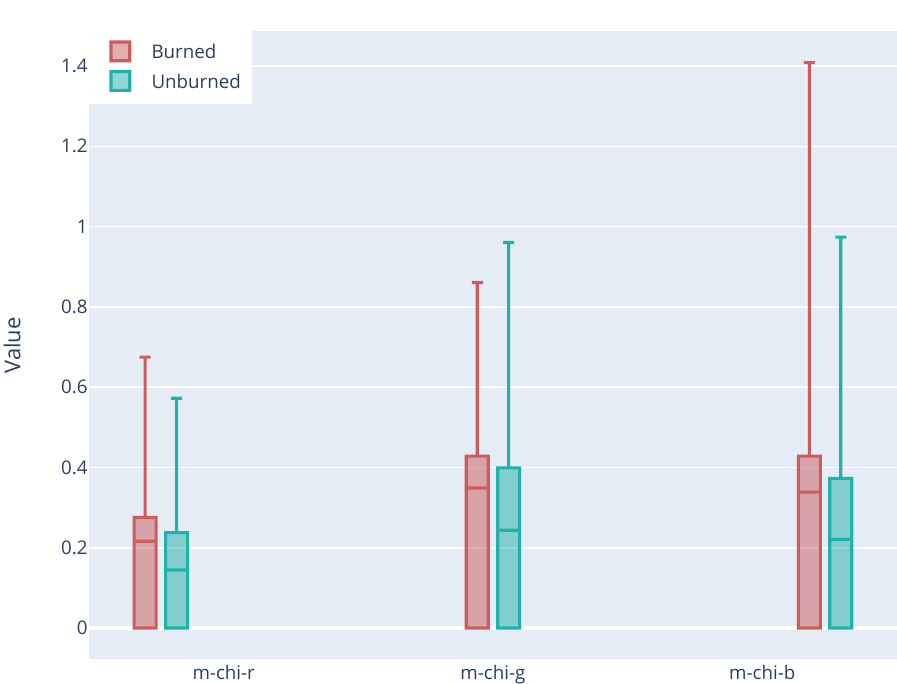}
\caption{The scatter plot of three channels of m-$\chi$ decomposition images for burned and unburned areas.}
\label{fig:scattermchi}
\end{figure}

The compact polarisation decomposition operation used to process the MLC products is provided through the European Space Agency's Sentinel Application Platform (SNAP). Common compact polarisation decomposition methods include m-$\chi$ decomposition, m-$\delta$ decomposition, h-$\alpha$ decomposition, RVOG decomposition and model-free 3-component Decomposition. As shown in Figure \ref{fig:decomp_comp}, all images—except those from the h-$\alpha$ decomposition—clearly visualize the burned area. However, there is little variation among the decomposition images generated by these four methods. For this research, the m-$\chi$ decomposition operator is selected. As shown in Equation \ref{decomp}, the R, G, and B elements of $m-\chi$ decomposition image can be derived from the parameters $m$ and $\chi$. 
\begin{equation}
    \label{decomp}
    \begin{split}
    m-\chi-R &= \sqrt{\frac{mg_0(1+sin2\chi)}{2}}\\
    m-\chi-G &= \sqrt{g_0(1-m)}\\
    m-\chi-B &= \sqrt{\frac{mg_0(1-sin2\chi)}{2}}
    \end{split}
\end{equation}
Barakat formulation is used to calculate the 2-D Barakat degree of polarization $m$ and the target characterization parameter $\chi$ as described in \cite{Dey2020TargetCA}. It requires the stroke parameters derived from the covariance matrix $C2$ as shown in Equation \ref{stroke}. $m$ and $\chi$ can be derived from Equation \ref{mchi}. The second column of Figure \ref{fig:inspect} shows the m-$\chi$ decomposition image, it can be observed that burned areas can be highlighted in the colour composite images. However, different colours of burned areas can be observed from the three study areas. Figure \ref{fig:scattermchi} provides how three bands of m-$\chi$ decomposition images change from unburned to burned areas. To generate the scatter plot, 5000 points are sampled from each image in the training dataset. It can be observed that the mean value of all three bands increases when the area is burned. Among the three bands, the m-$chi$-b band provides the largest difference between the mean value of the burned area and the unburned area, indicating the significance of this band. \par
\begin{equation}
\label{stroke}
    \textbf{S}=\begin{bmatrix}
S_0\\
S_1\\
S_2\\
S_3
\end{bmatrix}=
=\begin{bmatrix}
C_{11}+C{22}\\
C_{11}-C{22}\\
C_{12}+C{21}\\
\pm j(C_{12}-C_{21})
\end{bmatrix}
\end{equation}
\begin{equation}
    \label{mchi}
    \begin{split}
    m &= \frac{1}{g_0}\sqrt{g_0^2+g_1^2+g_2^2}\\
    sin2\chi &= -\frac{g_3}{mg_0}\\
    \end{split}
\end{equation}

\subsubsection{Compact Polarisation Vegetation Index}
The radar vegetation index of compact polarisation data is also derived from covariance matrix as shown in Equation \ref{c2matrix}. \cite{Mandal2020ARV} proposes the compact-pol radar vegetation index (CpRVI) which utilizes the geodesic distance (GD) projected on the unit sphere between two Kennaugh matrices which are commonly used to describe both coherent and incoherent scattering \cite{SCHMITT2015122}. The detailed derivation of geodesic distance can be found in \cite{Mandal2020ARV}. As shown in Equation \ref{cprvi}, the CpRVI can be derived using the fraction of the minimum backscattering and the maximum backscattering power, $g_0$ and $g_3$ are stroke parameters from Equation \ref{stroke}. As shown in the Third column of Figure \ref{fig:inspect}, the burned area of CpRVI image is darker than the unburned areas, which is caused by a lower vegetation because of the wildfire.
\begin{equation}
    \label{cprvi}
    CpRVI = (\frac{(min{\frac{g_0-g_3}{2}, \frac{g_0+g_3}{2}})}{max{(\frac{g_0-g_3}{2}, \frac{g_0+g_3}{2}})})^{2(\frac{3}{2}GD)}(1-\frac{3}{2}GD)
\end{equation}

\subsubsection{Discussions of the visualisation of the GRD, m-$\chi$ decomposition and CpRVI images}
From Figure \ref{fig:inspect}, it can be observed that all these data can be used to highlight the burned area. However, in the case of GRD images, the image in the second row is impacted by terrain effects, causing confusion between burned areas and mountainous regions. For G90273 and G90290, the CpRVI provides less contrast between burned and unburned areas when compared to GRD images and m-$\chi$ decomposition images. All GRD, m-$\chi$ and CpRVI images have excellent highlights of burned areas for EWF031 and RWF034. Different visualization effect of three data sources motivates the usage of a combination of these data in burned area mapping which is described in Section \ref{sec:methodology}.

\subsection{Pre-processing Pipeline}
\label{pre-proc}
\begin{figure}[hbt!]
  \centering
  \includegraphics[scale=0.55]{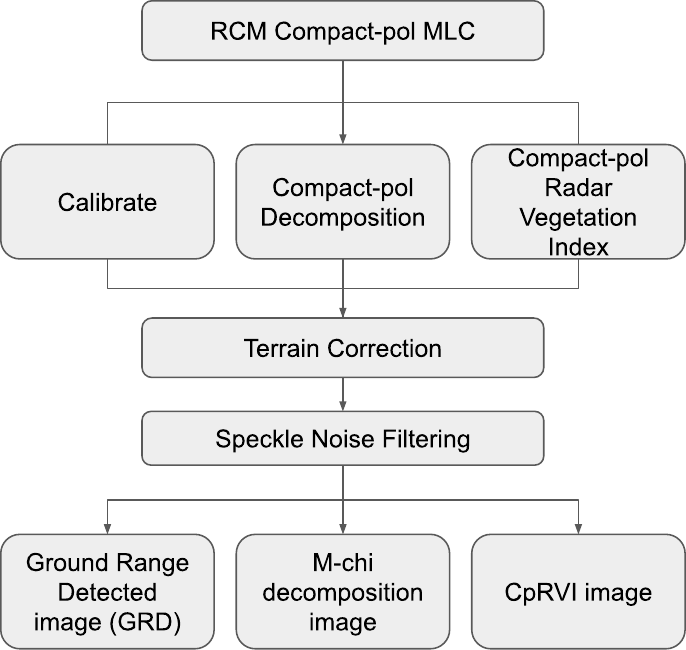}
\caption{The processing pipeline based on ESA SNAP that processes RCM MLC products to Ground Range Detection image, m-$\chi$ decomposition image and compact-pol radar vegetation index image.}
\label{fig:pipeline}
\end{figure}

The processing pipeline depicted in Figure \ref{fig:pipeline} is employed for processing the MLC data. For each of the above study areas in Figure \ref{fig:sa}, the RCM MLC data from May to October are selected to provide more varieties of images with different combinations of ground moisture, beam modes and orbits which helps the deep learning model to generalize better to new wildfires. This pipeline is based on the European Space Agency's Sentinel Application Platform (SNAP). MLC products are used in three separate branches to generate GRD, m-$\chi$ decomposition and CpRVI images. During the calibration process which is used to obtain the GRD images, raw digital numbers are converted to the sigma nought, which represents the backscatter coefficient in decimal format. Compact-pol decomposition and compact-pol radar vegetation index are derived from MLC products in the other two branches. Outputs of all three branches go through terrain correction and speckle noise filtering afterwards. In the terrain correction step, Digital Elevation Models (DEMs) are used to rectify distortions present in the raw satellite images. This correction accounts for the topographical variations of the landscape, ensuring that the data accurately reflects the true surface. Following terrain correction, the images undergo speckle noise filtering. This process is essential for eliminating speckle noise, which is a common issue in radar images caused by the coherence of the radar signal used by the sensor. \par
\begin{figure}[hbt!]
  \centering
  \includegraphics[scale=0.55]{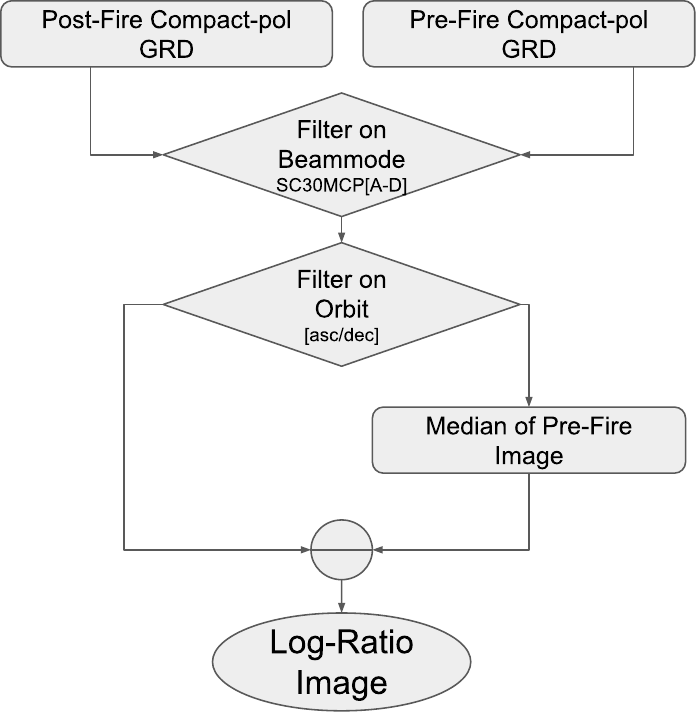}
\caption{The processing pipeline based on ESA SNAP that processes Ground Range Detection image to Log-ratio intensity image.}
\label{fig:logrt}
\end{figure}

To better visualize the burned area, the log-ratio intensity images are used as the input for the deep-learning model. As shown in Figure \ref{fig:logrt}, the pre-fire images and post-fire images are selected based on the same beam modes and the same orbit (ascending and descending) to ensure they share similar incidence angles. Then the median image of the pre-fire image collection is used as the pre-fire image. Finally, the log-ratio image is generated by using every image within the post-fire image collection to subtract the median of the pre-fire images. 

\section{Methodology}
\label{sec:methodology}
\subsection{Deep Learning-based Workflow}

\begin{figure}[hbt!]
  \centering
  \includegraphics[width=\linewidth]{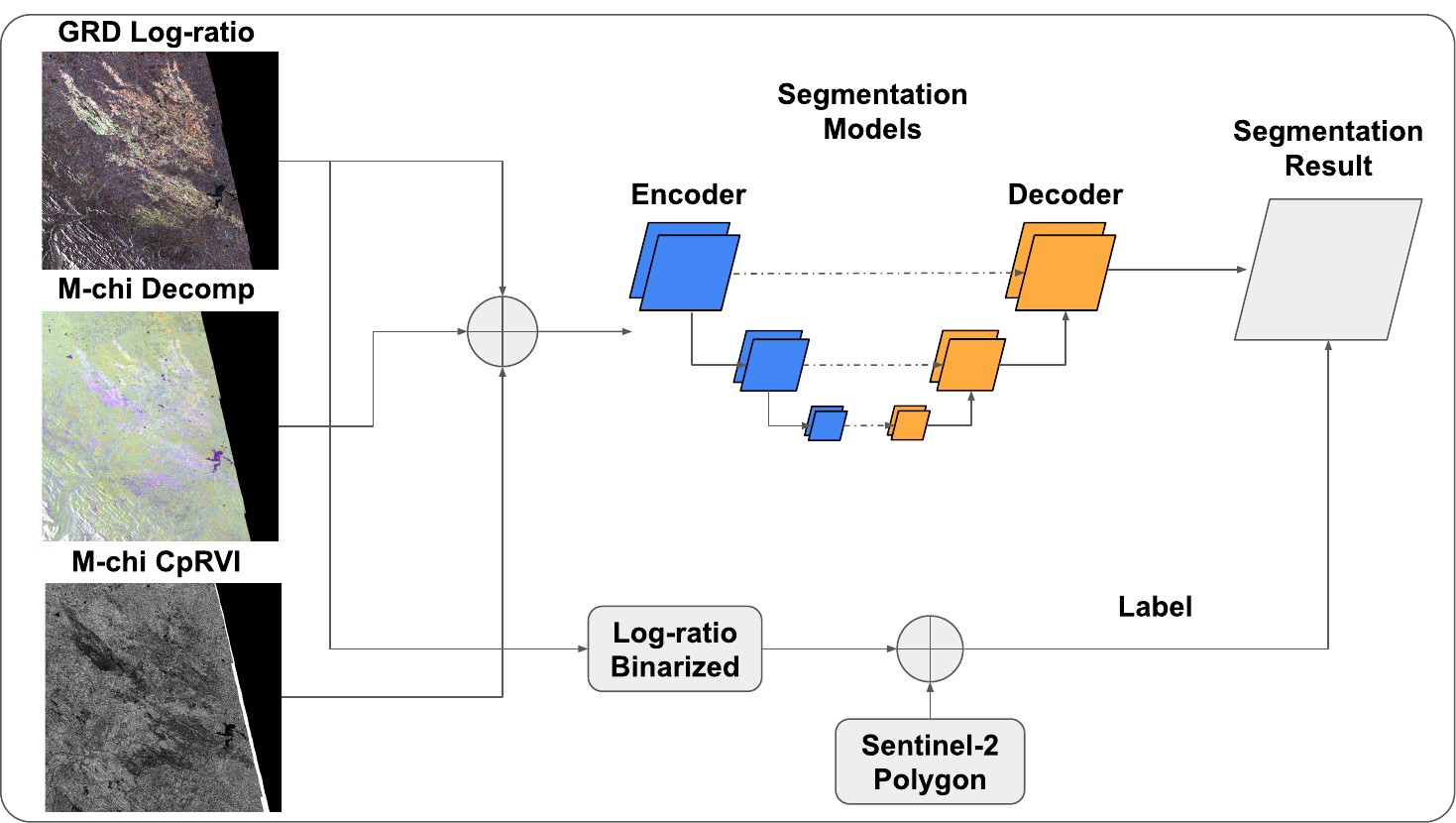}
\caption{Deep Learning Based Workflow for Burned Area Mapping using RCM Data.}
\label{fig:method}
\end{figure}

As shown in Figure \ref{fig:method}, the proposed deep learning workflow utilizes ConvNet-based and transformer-based deep learning segmentation models for burned area mapping. six different deep learning models are trained with the same dataset to assess how the choice of the models affects the burned area detection with RCM data. U-Net and Attention-U-Net are two ConvNet-based segmentation baseline models. U-Net is the classic baseline model for semantic segmentation which is widely applied to remote sensing applications. The skip-connection of U-Net can preserve high-level semantic information that helps with the burned area detection. Attention-U-Net is a stronger baseline that enables attention mechanism that enables learning with a focus on specific regions of the image. As for the Transformer-based models, UNETR and SwinUNETR-V2 are used as the strong baseline. Transformer becomes the major method for semantic segmentation after the emergence of Vision Transformer. Combining the architecture of U-Net and Vision Transformer, UNETR is proposed originally for medical image segmentation and shows promising performance. SwinUNETR is the latest segmentation model that adopts Swin-Transformer in the U-Net architecture. SwinUNETR-V2 improves the original SwinUNETR by introducing stagewise convolutions for features of each resolution level.\par
Since the training dataset comes from the images selected during and after the wildfire events. The training labels from optical satellites are not always available. Consequently, we use the manual thresholding method with log-ratio images to generate the pseudo labels from the RCM data for the burned area labels during the wildfire events. Each log-ratio image is binarized with a binarization threshold as shown in Equation \ref{binarization} to ensure the best visualization of the burned area. Among Equation \ref{binarization}, I refers to the input log-ratio image, $th$ is the threshold applied on the image which is set to 0.05, and M is the binary mask derived from the optical satellites images from Sentinel-2, Landsat-8 and Landsat-9 in British Columbia and Alberta. Wildfires often result in both increases and decreases in backscatter when compared to unaffected regions. This variation leads to extreme values, which can be effectively extracted by applying a threshold based on the standard deviation to the absolute value of the images. However, because of the speckle noise of the RCM data, the binarized image can be noisy. To better reduce the noise in the label, we utilize the polygons from the optical satellite data to filter out the speckle noise. For the post-fire burned area in the training, validation, and test dataset, binary masks of the post-fire optical images are utilized as labels. The binarized Normalized Burned Ratio (NBR) with a threshold of 0.1 is utilized to generate the binary mask, which is calculated using the normalized difference between the Short-wave Infrared and Near-infrared bands.\par

\begin{equation}
\label{binarization}
    I_{binarization} = (|I|>th) \cdot M
\end{equation}
\subsection{Setup and Evaluation Metrics}
All models are trained with Adam Optimizer with a learning rate of 3e-4. The batch size is set as 32 for both Transformer-based models and ConvNet-based models. And the models are trained for 100 epochs with the same random seed 42. The epoch with the best validation loss is selected to be tested on the test dataset. The loss function used in the training is the Dice Loss, which is used to maximize the intersection over the union between the output and the label. The Dice Loss has the formula as Equation \ref{dice}, TP means True Positive detection, FP means False positive detection and FN means False Negative detection.\par
\begin{equation}
\label{dice}
L_{dice}=1-\frac{2TP}{2TP+FP+FN}
\end{equation}
The performance of the models is evaluated with F1 Score and IoU Score as shown in Equation \ref{f1} and \ref{iou}. F1 Score is the reverse of the dice loss and IoU Score measures the intersection of the prediction and the label over the union of them. 
\begin{equation}
    \label{iou}
    \text{IoU} =\frac{TP}{FP+TP+FN}\\
    \end{equation}
    \begin{equation}
    \label{f1}
    F_1\text{ Score} =\frac{2TP}{2TP+(FP+FN)}
\end{equation}
\section{Results and Discussion}
\subsection{Quantitative Results}
\label{quantitative}

As shown in Table \ref{testset}, there are in total 7 study regions with 9 available images used in the test dataset. Two of these wildfire events, located in Quebec, are collected and evaluated together due to their close geolocation. Images from both ascending and descending orbits are utlized. Wildfire events are located in Quebec, Northwest Territory and British Columbia, Canada. There is one wildfire event in 2024 and 5 other wildfire events are from 2023. The labels used in the test dataset are generated using post-fire Sentinel-2 images. However, one of the British Columbia fires in 2024, there are old burned areas from 2023. For this fire event, we use the difference Normalized Burned Ratio to remove the burned area. The detection of the old burned area is also removed by using the polygon of the 2023 wildfires.\par
\begin{table}[!hbt]
\centering
\caption{Information about wildfire events used in the test dataset.}
\label{testset}
\begin{tblr}{
  cells = {c},
  cell{2}{1} = {r=5}{},
  cell{2}{2} = {r=3}{},
  cell{5}{2} = {r=2}{},
  cell{7}{1} = {r=2}{},
  hline{1,11} = {-}{0.08em},
  hline{2} = {-}{},
}
\textbf{Location}              & \textbf{Geolocation} & \textbf{Image Date} & \textbf{Orbit} \\
Quebec, Canada                 & (-75.49, 53.24)      & 2023-10-02          & Ascending      \\
                               &                      & 2023-10-05          & Decending      \\
                               &                      & 2023-10-05          & Ascending      \\
                               & (-76.30, 49.11)      & 2023-10-05          & Ascending      \\
                               &                      & 2023-10-27          & Ascending      \\
{Northwest Territory,\\Canada} & (-115.47, 62.77)     & 2023-10-01          & Decending      \\
                               & (-113.78, 62.80)     & 2023-10-10          & Decending      \\
{British Columbia,\\Canada}    & (-120.90, 58.35)     & 2024-05-27          & Ascending      \\
{British Columbia\\Canda}      & (-109.50, 60.30)     & 2023-10-04          & Decending      
\end{tblr}
\end{table}

As shown in \ref{ab:totalquant}, the quantitative results of six study regions are provided. The average F1 Score and IoU Score for all available images of each wildfire event are provided. Transformer-based model UNETR provides the best performance on average and ConvNet-based model Attention-U-Net provides similar accuracy compared to UNETR. But noticeably, in terms of generalizability, UNETR and Attention-U-Net provides a worse F1 Score in one of the study regions from Northwest Territory compared to U-Net. SwinUNETR shows good results for wildfires in Quebec and British Columbia but performs poorly in wildfires in the Northwest Territory.\par
\begin{figure*}[!hbt]
  \centering
  \includegraphics[width=.8\linewidth]{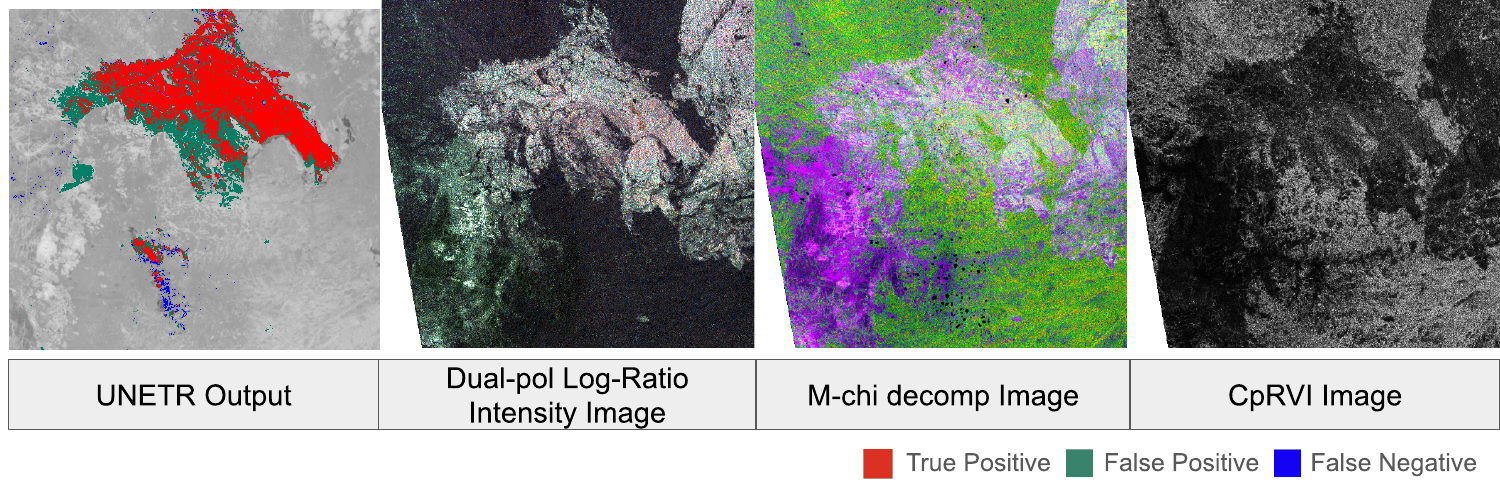}
\caption{Qualitative results of Wildfire near Kahntah (-120.90, 58.35), British Columbia, Canada. The output of the deep learning model is overlaid on top of the normalized burned ratio image of Sentinel-2.}
\label{fig:kahntah}
\end{figure*}
\subsection{Qualitative Results}

\begin{figure*}[!hbt]
  \centering
  \includegraphics[width=\linewidth]{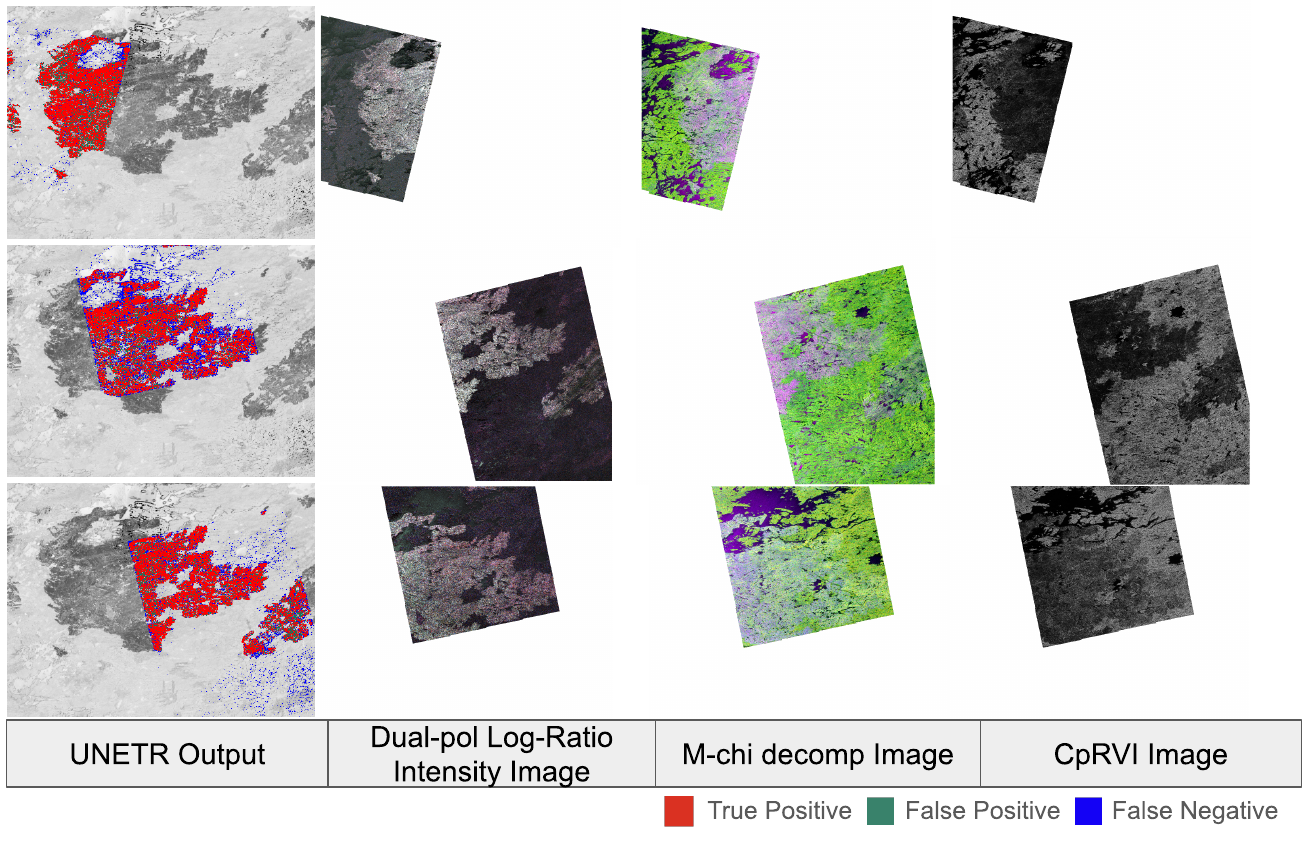}
\caption{Qualitative results of wildfire event in Quebec (-75.49, 53.24), Canada. The output of the deep learning model is overlaid on top of the normalized burned ratio image of Sentinel-2.}
\label{fig:quebec}
\end{figure*}

\begin{table*}[!hbt]
\centering
\caption{Quantitative results of six deep learning segmentation models over different locations. All results are based on models trained with log-ratio, compact-pol decomposition and CpRVI data.}
\label{ab:totalquant}
\scalebox{0.8}{
\begin{tblr}{
  cells = {c},
  cell{1}{1} = {r=2}{},
  cell{1}{2} = {r=2}{},
  cell{1}{3} = {c=2}{},
  cell{1}{5} = {c=2}{},
  cell{1}{7} = {c=2}{},
  cell{1}{9} = {c=2}{},
  cell{3}{1} = {r=2}{},
  cell{5}{1} = {r=2}{},
  cell{7}{1} = {r=2}{},
  hline{1,3,10} = {-}{},
  hline{2} = {3-10}{},
  hline{5,7,9} = {2-10}{},
}
\textbf{Location}                     & \textbf{\textbf{Coordinates}} & \textbf{\textbf{U-Net}} &              & \textbf{\textbf{Attention-U-Net}} &              & \textbf{\textbf{SwinUNETR}} &              & \textbf{\textbf{UNETR}} &              \\
                                      &                               & \textbf{F1}             & \textbf{IoU} & \textbf{F1}                       & \textbf{IoU} & \textbf{F1}                 & \textbf{IoU} & \textbf{F1}             & \textbf{IoU} \\
\textbf{\textbf{British Colombia}}    & (-109.50, 60.30)              & 0.695                   & 0.532        & 0.733                             & 0.579        & 0.681                       & 0.517        & 0.670                   & 0.504        \\
                                      & (-120.90, 58.35)              & 0.692                   & 0.529        & 0.765                             & 0.619        & 0.763                       & 0.616        & 0.755                   & 0.607        \\
\textbf{\textbf{Quebec}}              & (-75.49, 53.24)               & 0.685                   & 0.525        & 0.780                             & 0.640        & 0.761                       & 0.619        & 0.790                   & 0.654        \\
                                      & (-76.30, 49.11)               & 0.667                   & 0.501        & 0.721                             & 0.564        & 0.712                       & 0.553        & 0.706                   & 0.546        \\
\textbf{\textbf{Northwest Territory}} & (-115.47, 62.77)              & 0.706                   & 0.545        & 0.693                             & 0.530        & 0.588                       & 0.416        & 0.690                   & 0.527        \\
                                      & (-113.78, 62.80)              & 0.715                   & 0.556        & 0.440                             & 0.282        & 0.276                       & 0.160        & 0.566                   & 0.395        \\
\textbf{Average}                      &                               & 0.688                   & 0.527        & 0.713                             & 0.562        & 0.668                       & 0.519        & 0.718                   & 0.565        
\end{tblr}}
\end{table*}
\begin{figure*}[h]
  \centering
  \includegraphics[width=\linewidth]{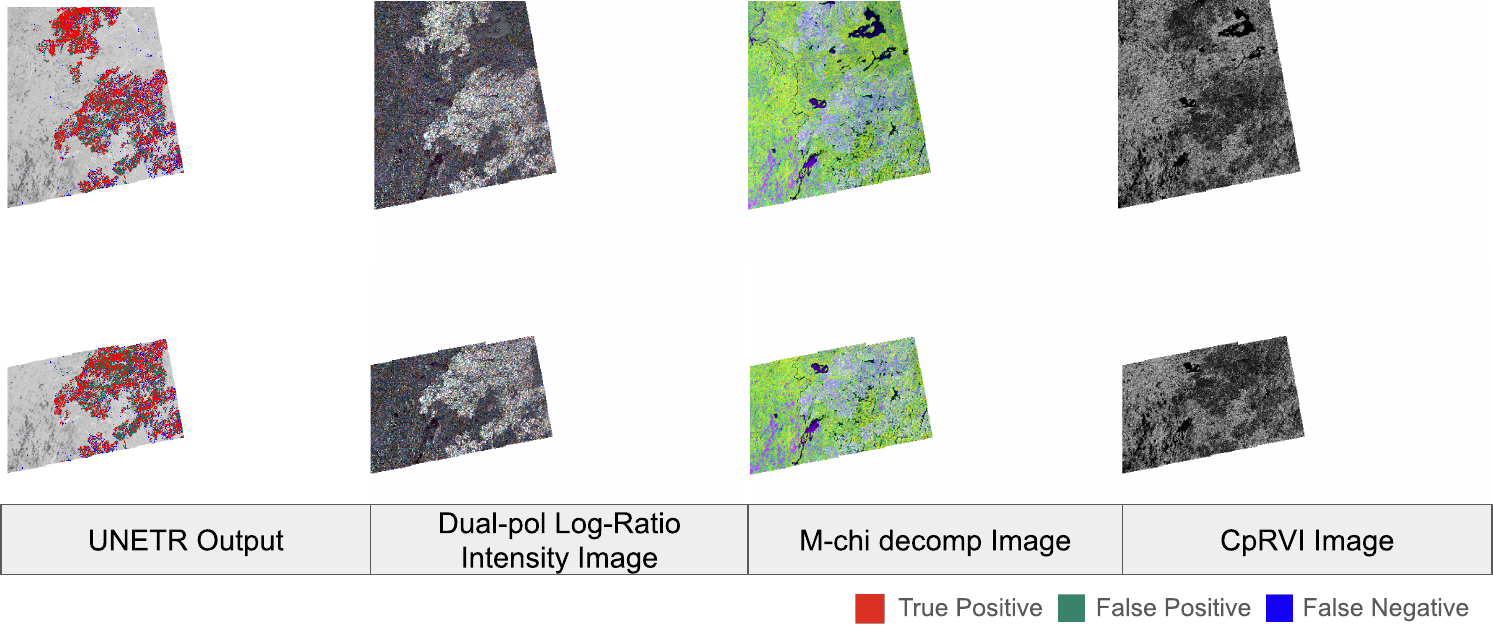}
\caption{Qualitative results of wildfire event in Quebec (-76.30, 49.11), Canada. The output of the deep learning model is overlaid on top of the normalized burned ratio image of Sentinel-2.}
\label{fig:quebec2}
\end{figure*}
As shown in Figure \ref{fig:kahntah}, \ref{fig:quebec} and \ref{fig:quebec2}, qualitative results of wildfires in British Columbia and Quebec, Canada are presented. The wildfire in British Columbia happened near the Kahntah Reserve, resulting in an evacuation order and a state of emergency. The wildfire event happened in July 2024 and caused severe damage to the forest. As shown in Figure \ref{fig:kahntah}, the output of UNETR can efficiently detect the burned area with some overestimation near the border of the wildfire. All log-ratio, m-chi decomposition and CpRVI images can efficiently distinguish the burned and unburned areas. Among these three data sources, Log-ratio and CpRVI provides the best contrast. Quantitatively, using all three data sources provides an F1 Score of 0.755 and an IoU Score of 0.607. Using only the log-ratio images fails to detect the burned area completely while using Compact-pol Decomposition (CpDcomp) plus CpRVI achieves an F1 Score of 0.694 and an IoU Score of 0.531.\par

As for the wildfires that happened in Quebec, Canada, in Figure \ref{fig:quebec}, the deep learning output can all detect the burned areas with some underestimation caused by the artefacts of the optical labels because of the cloud. Compact-pol decomposition image can visualize the burned area in the first two images, but the effect is less significant in the third image. CpRVI highlights most of the burned area but in the third image, there is a large dark region that can be easily confused with the burned area caused by the lake. The log-ratio images provide a strong contrast between burned and unburned areas in the first and third images. However, for the image in the second row, the burned areas on the right appear darker than those on the left, making it challenging for the deep-learning model to detect them accurately. For the other study area in Quebec shown in \ref{fig:quebec2}, the deep learning results also clearly detect the burned area. However, the compact-pol decomposition images are less obvious compared to the first study area. The log-ratio images and CpRVI images both provide a good contrast between burned and unburned areas. In summary, based on the qualitative results and visualization of the input bands, it is evident that these three bands complement each other, which enhances burned area detection. This observation also supports the conclusions drawn from the Ablation Study in Section \ref{ablation}.
\subsection{Ablation Study on the Effect of Compact-pol Decomposition and CpRVI}
\label{ablation}
To investigate the effect of compact-pol decomposition and CpRVI, models area trained with different input bands. As shown in Table \ref{tab:avgmodel}, the average F1 Score and IoU Score of six different models for four settings of input bands including using solely Log-ratio images, only Compact-pol Decomposition (CpDcomp) image, only Compact-pol Decomposition image plus CpRVI, and using all bands available are provided. It can be observed overall, that using log-ratio images and other two data sources provides significantly better results in detecting the burned area (F1 Score:0.697, IoU Score: 0,543).\par
As shown in Table \ref{ab:input}, results of UNETR with three settings in each study area are provided. The quantitative results reported in Table \ref{ab:input} are based on results of UNETR trained with three different settings. Overall the UNETR trained with all bands provides the best F1 Scores (0.718) and IoU scores (0.565). It is significantly improved from UNETR trained with Log-ratio images only which provides 0.684 and 0.557 as the F1 Score and IoU Score. For the UNETR trained with only post-fire compact-pol decomposition image and CpRVI, the model provides much lower F1 Score and IoU Score compared to the models trained with log-ratio images. This indicates that using only compact-pol decomposition images and CpRVI cannot efficiently detect burned areas. Moreover, the metrics of the model trained solely with compact-pol decomposition images indicate that CpRVI helps to improve detection accuracy. Especially, UNETR trained with only Compact-pol decomposition images completely misses the detection of some study regions, which indicates low generalization ability.\par
When comparing the quantitative results of each fire event, it can be observed that using only log-ratio images achieves better detection results in the first wildfire events of Quebec fires and the second wildfire events of Northwest Territory events than using all data sources. However, failing to detect the burned area of the second wildfire event in British Columbia makes the quantitative results worse than using all data sources. It also highlights the importance of compact-pol decomposition and CpRVI images as complementary data source to the log-ratio images for burned area mapping.

\begin{table*}[t]
\centering
\caption{Ablation study for various settings of input bands for each study area. All results are based on UNETR.}
\label{ab:input}
\scalebox{0.8}{
\begin{tblr}{
  cells = {c},
  cell{1}{1} = {r=2}{},
  cell{1}{2} = {r=2}{},
  cell{1}{3} = {c=2}{},
  cell{1}{5} = {c=2}{},
  cell{1}{7} = {c=2}{},
  cell{1}{9} = {c=2}{},
  cell{3}{1} = {r=2}{},
  cell{5}{1} = {r=2}{},
  cell{7}{1} = {r=2}{},
  hline{1,3,5,7,9-10} = {-}{},
  hline{2} = {3-10}{},
}
\textbf{Location}                                                                               & \textbf{Coordinates} & \textbf{Log-Ratio } &              & {\textbf{\textbf{Decomp +~}}\\\textbf{\textbf{CpRVI}}} &              & {\textbf{\textbf{Decomp}}} &              & {\textbf{Log-Ratio +~}\\\textbf{Decomp +~}\\\textbf{CpRVI}} &                \\
                                                                                                &                      & \textbf{F1}         & \textbf{IoU} & \textbf{F1}                                            & \textbf{IoU} & \textbf{F1}                                            & \textbf{IoU} & \textbf{F1}                                                 & \textbf{IoU}   \\
{\textbf{\textbf{\textbf{\textbf{British }}}}\\\textbf{\textbf{\textbf{\textbf{Columbia}}}}}    & (-109.50, 60.30)     & 0.717               & 0.559        & 0.622                                                  & 0.451        & 0.586                                                  & 0.414        & 0.670                                                       & 0.504          \\
                                                                                                & (-120.90, 58.35)     & 0.023               & 0.011        & 0.694                                                  & 0.531        & 0.001                                                  & 0.000        & 0.755                                                       & 0.607          \\
\textbf{\textbf{\textbf{\textbf{Quebec}}}}                                                      & (-75.49, 53.24)      & 0.810               & 0.681        & 0.656                                                  & 0.495        & 0.699                                                  & 0.539        & 0.790                                                       & 0.654          \\
                                                                                                & (-76.30, 49.11)      & 0.687               & 0.523        & 0.655                                                  & 0.487        & 0.638                                                  & 0.468        & 0.706                                                       & 0.546          \\
{\textbf{\textbf{\textbf{\textbf{Northwest~}}}}\\\textbf{\textbf{\textbf{\textbf{Territory}}}}} & (-115.47, 62.77)     & 0.695               & 0.533        & 0.661                                                  & 0.494        & 0.638                                                  & 0.469        & 0.690                                                       & 0.527          \\
                                                                                                & (-113.78, 62.80)     & 0.846               & 0.733        & 0.067                                                  & 0.035        & 0.004                                                  & 0.002        & 0.566                                                       & 0.395          \\
\textbf{Average}                                                                                &                      & 0.684               & 0.557        & 0.564                                                  & 0.415        & 0.428                                                  & 0.315        & \textbf{0.718}                                              & \textbf{0.565} 
\end{tblr}}
\end{table*}
\begin{table}[!hbt]
\centering
\caption{Average F1 Score and IoU Score of six deep learning models for four settings of input bands.}
\label{tab:avgmodel}
\begin{tblr}{
  column{2} = {c},
  column{3} = {c},
  hline{1-2,6} = {-}{},
}
\textbf{Input Bands}         & \textbf{F1 Score} & \textbf{IoU Score} \\
Log-ratio                    & 0.653             & 0.520              \\
CpRVI + CpDecomp             & 0.455             & 0.328              \\
CpDecomp                     & 0.419             & 0.297              \\
Log-ratio + CpRVI + CpDecomp & \textbf{0.697}    & \textbf{0.543}     
\end{tblr}
\end{table}
\subsection{Discussion}
As discussed in the quantitative and qualitative results, compact-pol decomposition images and CpRVI images can help to detect burned areas better compared to only using log-ratio images. However, from the results in Section \ref{ablation}, it can be observed that the models trained with only the compact-pol decomposition and CpRVI images do not provide promising results, even compared with the models trained with only log-ratio images. It indicates that the backscatter images are still the most useful input for burned area detection. Compact-pol decomposition and CpRVI are complementary to log-ratio images which helps different models to achieve better performance. \par
When cross-comparing the metrics between six different deep learning models in Table \ref{ab:totalquant}, it can be observed that the difference between the best-performing model and the worst-performing model is below 0.1, while as shown in \ref{tab:avgmodel}, the performance gap between different inputs is close to 0.3. It highlights the importance of feature engineering of SAR images to create more meaningful features for burned area detection.\par
\section{Conclusion}
In this research, we introduce a pre-processing pipeline for RADARSAT Constellation Mission compact polarization SAR data to generate log-ratio, compact-pol decomposition and compact-pol radar vegetation index from RCM’s MLC products. Subsequently, a wildfire dataset is created using these RCM log-ratio images based on Canadian wildfires in 2023 and tested on Canadian wildfire events in 2023 and 2024. Then we evaluated six different deep learning models with three settings of input by using only log-ratio, using only compact-pol decomposition plus CpRVI and using all three data source. Our results demonstrate that the compact-pol decomposition images and Compact-pol Radar vegetation index images complement the log-ratio images effectively for burned area detection. By using compact-pol decomposition and CpRVI images, the best-performing deep learning model achieves an F1 Score of 0.718 and an IoU Score of 0.565 compared to the models trained only with log-ratio images (F1 Score: 0.653 and IoU Score: 0.520).

\section*{Acknowledgement}
The research is part of the project ‘EO-AI4Global Change’ funded by Digital Futures, and the project ‘SAR4Wildfire’ funded by Formas, the Swedish research council for sustainable development. Thanks to the Canadian Space Agency for providing the RCM data.

\bibliographystyle{elsarticle-harv.bst} 
\bibliography{refs}




\end{document}